\documentclass[10pt,twocolumn,letterpaper]{article}

 \usepackage{iccv}              
\usepackage{multirow} 
\usepackage{amsmath}
\usepackage[accsupp]{axessibility} 

\definecolor{iccvblue}{rgb}{0.21,0.49,0.74}
\usepackage[pagebackref,breaklinks,colorlinks,allcolors=iccvblue]{hyperref}

\def\paperID{298} 
\def\confName{ICCV}
\def\confYear{2025}

\title{\textit{LongAnimation}: Long Animation Generation with Dynamic Global-Local Memory}

\author{
Nan Chen\textsuperscript{1}, Mengqi Huang\textsuperscript{1}, Yihao Meng\textsuperscript{2}, Zhendong Mao\textsuperscript{1†}\\
\textsuperscript{1}University of Science and Technology of China\\
\textsuperscript{2}Hong Kong University of Science and Technology\\
{\tt\small \{chen\_nan,huangmq\}@mail.ustc.edu.cn, ymengas@connect.ust.hk, \{zdmao\}@ustc.edu.cn}
}

\begin{document}
\maketitle

\renewcommand{\thefootnote}{\fnsymbol{footnote}}
\footnotetext[2]{Zhendong Mao is the corresponding author.}

\begin{abstract}
Animation colorization is a crucial part of real animation industry production. Long animation colorization has high labor costs. Therefore,  automated long animation colorization based on the video generation model has significant research value. Existing studies are limited to short-term colorization. These studies adopt a local paradigm,  fusing overlapping features to achieve smooth transitions between local segments. However,  the local paradigm neglects global information,  failing to maintain long-term color consistency. In this study,   we argue that ideal long-term color consistency can be achieved through a \textbf{dynamic global-local paradigm},  i.e.,   dynamically extracting 
global color consistent features relevant to the current generation. Specifically,  we propose \textbf{\textit{LongAnimation}},  a novel framework,   which mainly includes a SketchDiT,  a Dynamic Global-Local Memory (DGLM),  and a Color Consistency Reward. The SketchDiT captures hybrid reference features to support the DGLM module. The DGLM module employs a long video understanding model to dynamically compress global historical features and adaptively fuse them with the current generation features. To refine the color consistency,  we introduce a Color Consistency Reward. During inference,  we propose a color consistency fusion to smooth the video segment transition. Extensive experiments on both short-term (14 frames) and long-term (average 500 frames) animations show the effectiveness of \textit{LongAnimation} in maintaining short-term and long-term color consistency for open-domain animation colorization task. The code can be found at \href{https://cn-makers.github.io/long_animation_web/}{LongAnimation}.

\end{abstract}

\section{Introduction}
\label{sec:intro}

\begin{figure}[!t]
\centering
\includegraphics[width=1\linewidth]{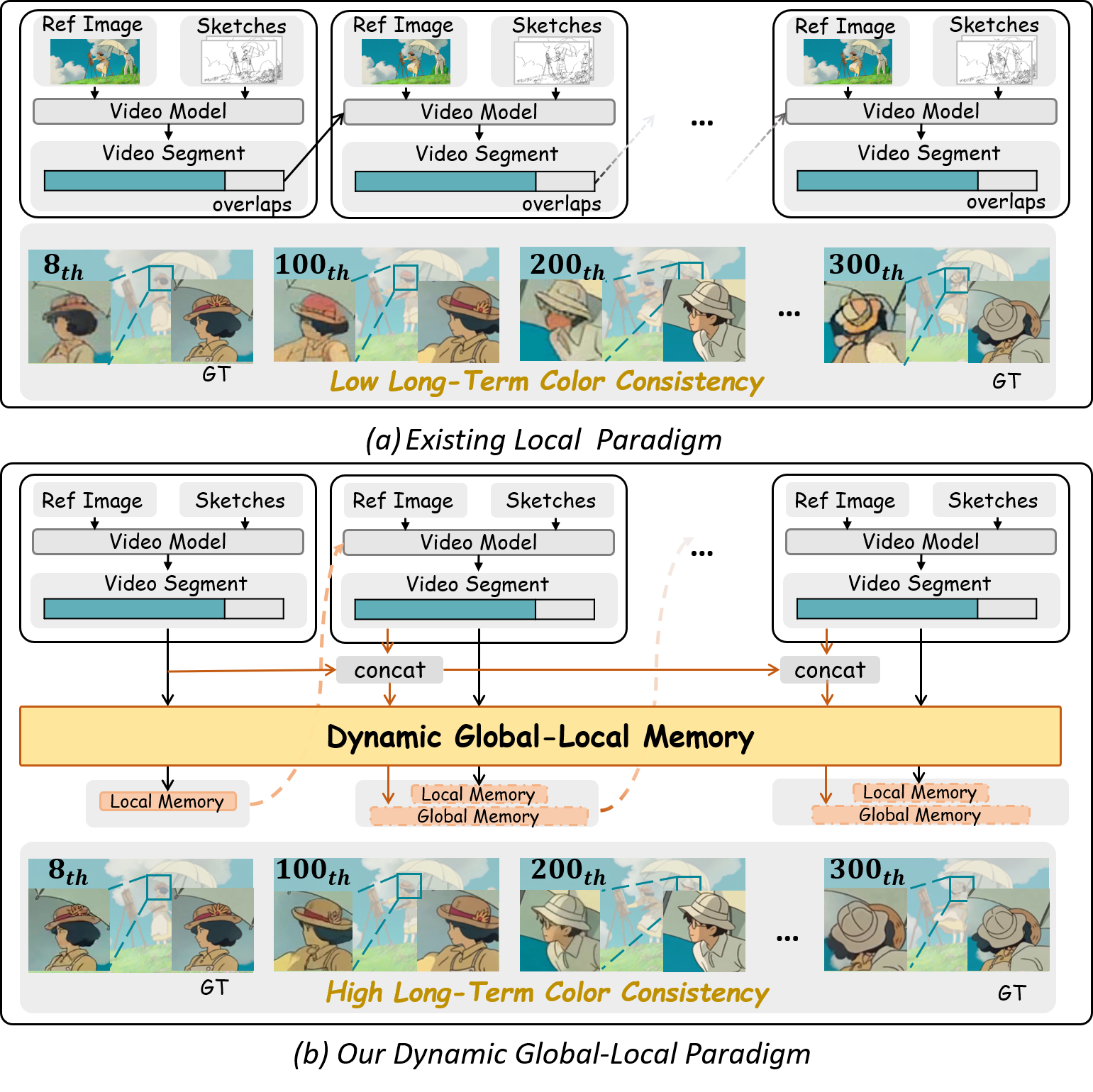}
\caption{Comparison with existing paradigm. (a) Existing studies achieve local color consistency by fusing the overlaps of adjacent video segments,  suffering low long-term color consistency. (b) Our dynamic global-local paradigm dynamically extracts color features of global historical segments as global memory and the color features of the latest generated segment as local memory,  achieving high long-term color consistency. All segments are generated from one same reference image.
}
\label{fig_intro}
\end{figure}

Animation is one of the most entertaining and aesthetic forms in the video industry. Animation colorization is a crucial part of real animation industry production. The real animation industry first draws and colors one key frame (not just the first frame,  but also the frame in the sequence). Then,  sketch sequences around key frame,  especially long sequences up to $10 \sim 30$ seconds (300 to 1000 frames),  are colored based on the key frame. Coloring such long sequences is extremely labor-intensive,  which makes the research of automated long animation coloring highly significant. The primary goal of long animation coloring is maintaining long-term color consistency,  which means the same object should have consistent colors in all frames.

Existing methods mainly explore short-term (within 100 frames) animation colorization. Early work \cite{yu2024animation} colors each frame based on image models and splices them into videos with weak color consistency. Recent studies \cite{xing2024tooncrafter, yang2025layeranimate, meng2024anidoc}, based on video models, explore various control on a fixed segment (\eg,  16 frames),  such as frame interpolation \cite{xing2024tooncrafter},  single-layer coloring \cite{yang2025layeranimate},  and character-based coloring \cite{meng2024anidoc}. These methods do not consider longer animation coloring. LVCD \cite{huang2024lvcd},  based on the Unet video model,  further attempts to use local paradigm to extend colorization up to 100 frames,  that is,  fusing the overlapping features of adjacent segments,  as shown in \cref{fig_intro} (a). In summary,  existing methods are either limited to fixed short-term coloring or use local paradigm to achieve limited coloring extension.

However,  existing methods have inherent defects,   \emph{i.e.},  neglecting the global color relationship,  which makes it difficult to maintain color consistency over longer animations. Autoregressively using the last frame of the previous generation as the reference image for the next segment causes noise error accumulation in long videos,  while using only one fixed reference image to control all segments ignores the relationship between segments. Based on a fixed reference image,  the local paradigm fuses the overlapping frame features of adjacent segments to partially constrain the consistency of adjacent segments rather than the color consistency between all segments. Additionally,  some animations involve large movements,  which makes keeping long-term consistency with local paradigm more challenging. As a result,  erratic color details appear in long animations (\eg,  the girl's yellow hat turns red in the 100th frame,  and the boy's hat changes color in the 300th frame),  as shown in \cref{fig_intro}(a).

We argue that an ideal long animation coloring performance could be achieved by the \textbf{dynamic global-local} paradigm,  which aims to achieve long-term (\eg,  500 frames) consistency by dynamically fusing global and local color features,  as shown in \cref{fig_intro}(b). The history animations contain color consistency features and redundant features. Global memory can dynamically extract color consistency features,  achieving global color consistency constraint. Additionally,  global memory can model low-frequency animation changes to achieve more stable colorization,  reducing color flickering,  as shown in \cref{visualization_1}. Local memory is used for short-term segments to ensure smooth local transitions. By the joint memory of global and local segments,  the dynamic global-local paradigm can effectively improve the long-term color consistency of animations (\eg,  the consistent color of the girl and boy's hat in the \cref{fig_intro}(b)).

In this paper,  we propose a novel DiT-based long animation coloring framework,  \textbf{\textit{LongAnimation}},  which achieves long-term color consistency by dynamically extracting color features related to current generation from global generation information. \textit{LongAnimation} mainly includes three key components: (1) SketchDiT,  which aims to efficiently extract reference features to support the subsequent dynamic memory mechanism; (2) Dynamic Global-Local Memory (DGLM) mechanism,  which aims to extract long-term consistency features related to the current generation from historical animations,  and (3) Color Consistency Reward,  which aims to refine the color consistency. Specifically,  we first propose the SketchDiT architecture for the animation colorization task,  which achieves efficient fusion of hybrid features of reference images,  sketches and text based on DiT-based video model. The innovative DGLM mechanism,  which first introduces the Long Video Understanding model to dynamically compress generated animations,  adaptively extracts global color consistent features. Furthermore,  we propose the Color Consistency Reward to further refine the model's coloring ability by aligning the low-frequency features extracted from the Long Video Understand model of the generated animation and reference animation. During inference,  we propose color consistency fusion to achieve smooth transitions,  which utilizes latent fusion in the later denoising stage. \textit{LongAnimation} can stably colorize long animations averaging \textbf{500} frames,  at least 5 times more than past methods \cite{huang2024lvcd, xing2024tooncrafter, meng2024anidoc} could achieve.

Our main contributions are summarized as follows:
\begin{itemize}
    \item \textbf{Concepts.} We propose a novel dynamic global-local paradigm,  which aims to achieve accurate long-term animation colorization by dynamically extracting global and local consistent color features.
    \item  \textbf{Technology.}  We propose \textit{LongAnimation},  which designs a novel DGLM module based on dynamic compression of long video understanding model to adaptively extract historical color features related to the current generation. To facilitate DGLM module,  we design a novel SketchDiT to efficiently control the DiT-based video model.
    \item  \textbf{Experiments.} \textit{LongAnimation} significantly outperforms previous advanced models on the open-domain long animation coloring task,  improving short-term and long-term performance by 35.1\% and 49.1\% respectively on the widely used metric Frechet Video Distance (FVD).
\end{itemize}

\section{Related Work}
\label{sec:related_work}
\subsection{Long Video Generation}
Visual generation has gradually evolved from controllable T2I \cite{lin2025realgeneral,jia2025d,zhang2025magiccolor} to controllable T2V \cite{wang2025dualreal,ma2024followyouremoji,ma2025followyourmotion,ma2025followyourclick,ma2024followpose}, in which a key issue is long video generation. Training-free studies \cite{qiu2023freenoise, lu2024freelong} extend the length by window fusion or frequency control. DitCTrl \cite{cai2024ditctrl} follows window fusion  in DiT \cite{peebles2023scalable} architecture. Training-based studies \cite{ren2024consisti2v, henschel2024streamingt2v, zeng2024make} usually rely on the most recent segment to generate the next segment. The most recent study \cite{hong2024slowfast} dynamically updates historical segment features by fine-tuning LoRA during inference. However,  long video generation with finer-grained control (\eg,  sketch,  ID) is more challenging as it requires more precise long-term consistency of control conditions. Some studies \cite{huang2024lvcd, zhang2024mimicmotion} introduce local fusion but rarely consider global consistency. \textit{LongAnimation} innovatively uses Long Video Understanding model to extract global temporal features for long controllable video generation.

\begin{figure*}[!t]
\centering
\includegraphics[width=1\linewidth]{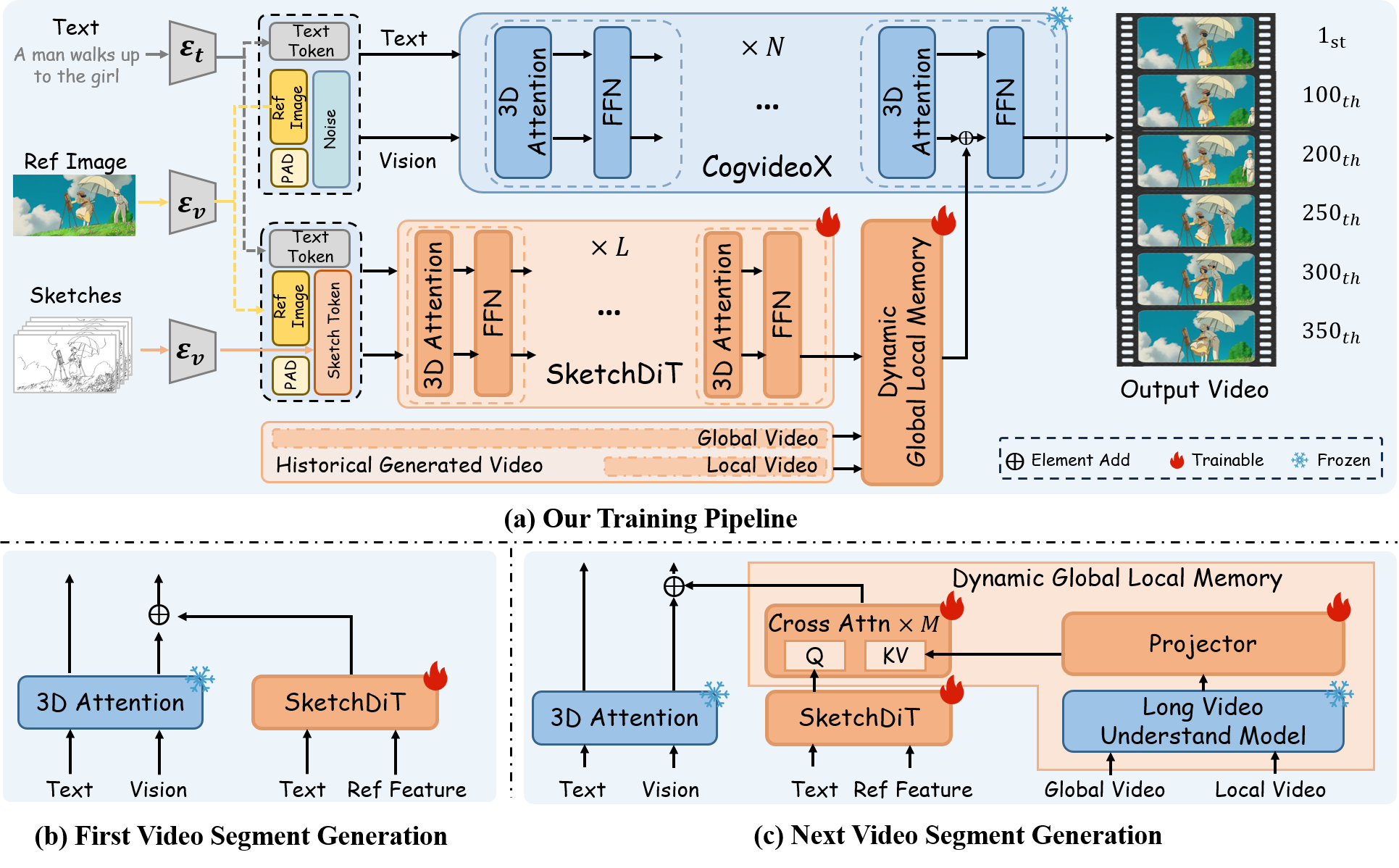}
\caption{Overview of the  \textit{LongAnimation}. (a) During training,  the reference information is fed into the CogvideoX \cite{yang2024cogvideox} and SketchDiT, respectively, for efficient extraction of hybrid reference features. These reference features are then fused with the historical information in Dynamic Global-Local Memory (DGLM) for consistency generation. (b) For the first segment generation,  the reference features are fed into SketchDiT and then directly sent to the video model. (c) For the subsequent segment generation,  DGLM dynamically extracts historical features,  which are adaptively fused with current reference features from the SketchDiT before being fed into the video model.} 
\label{fig_model_architecture}
\end{figure*}

\subsection{Long Video Understanding}
The multimodal community has explored many methods for long video understanding (LVU). Early studies \cite{lin2023video, li2024mvbench, hong2024cogvlm2} processes videos through sparse sampling (\eg,  16 and 24 frames). However,  sparse sampling discards numerous frames,  causing biased understanding. To achieve better understanding,  some studies \cite{wang2024longllava, shu2024video, wang2024retake, li2024llama, weng2024longvlm, fei2024video} compress video features to feed more frames into LVU models. Some studies \cite{li2024llama, weng2024longvlm, fei2024video} reduce the token numbers per frame or use token merging to compress features. Recent Video-XL \cite{shu2024video} and ReTAKE \cite{wang2024retake} use KV cache to compress visual input,  achieving fine-grained visual perception.

\subsection{Reference-based Line Art Video Coloring}
Reference-based line art video coloring focuses on injecting reference frame color (not just the first frame) into the animations. Early study ACOF \cite{yu2024animation} proposes transferring color for each frame by optical flow and splicing them into videos. Recent studies mainly use the sketch to generate consistent videos based on the Video Diffusion Model. ToonCrafter \cite{xing2024tooncrafter} proposes an animation interpolation diffusion model,  which requires the first and last reference frames. LVCD \cite{huang2024lvcd} introduces video ControlNet to control the motion,   using local fusion to improve the consistency of segment splicing. LayerAnimate \cite{yang2025layeranimate} proposes a Layer ControlNet to achieve single object control. AniDoc \cite{meng2024anidoc} can generate character animations given a character image without background. Recent studies \cite{huang2024lvcd,xing2024tooncrafter,meng2024anidoc,yang2025layeranimate} focus on generating short-term animations (\eg,  within 100 frames) based on the UNet video model. However, long animation shots often last $10 \sim 30$ seconds (300 to 1000 frames) in real scenarios. We propose \textit{LongAnimation} to achieve efficient long-term coloring on the DiT architecture.

\section{Methodology}
\label{sec:methodology}
The pipeline of \textit{LongAnimation} is depicted in \cref{fig_model_architecture},  which mainly consists of three parts: SketchDiT,  Dynamic Global-Local Memory and Color Consistency Reward. During training,  given a reference image \(I\),  \(F\) frames of sketches \(\{S_f\}_{f=1}^F\),   and corresponding text descriptions,  \textit{LongAnimation} extracts hybrid reference features of these inputs through SketchDiT,  which is designed to facilitate the subsequent Dynamic Global-Local Memory mechanism. When generating subsequent segments,  Dynamic Global-Local Memory dynamically compresses historical features,  which adaptively extracts global consistency features related to the current hybrid features obtained from SketchDiT,  ultimately generating \(F\) frames of long-consistency animation \(\{I_f\}_{f=1}^F\). Color Consistency Reward is used to refine  long-term color consistency during training while color consistency fusion is used to smooth transition during inference. In this section,  we will first introduce preliminaries. Then,  we will describe the SketchDiT,  Dynamic Global-Local Memory and Color Consistency Reward. Next,  we will introduce the color consistency fusion. 

\subsection{Preliminaries} 
\textbf{Video Diffusion Transformer.}   Our work is based on CogvideoX \cite{yang2024cogvideox},  one of the excellent performing DiT-based video generation models \cite{yang2024cogvideox, hacohen2024ltx, kong2024hunyuanvideo}. The text encoder $\boldsymbol{\mathcal{E}}_t(\cdot)$  first encodes the text into \(\boldsymbol{c}_t\). Next,  the reference image and video are encoded by the 3D VAE encoder $\boldsymbol{\mathcal{E}}_v(\cdot)$ into reference image token \(\boldsymbol{c}_i\) and video token \(\boldsymbol{z}\). Then,  \(\boldsymbol{c}_i\) is padded and concatenated with \(\boldsymbol{z}\) along the channel dimension,  followed by concatenation with \(\boldsymbol{c}_t\) along the sequence dimension. Finally,  the combined sequence is fed into the DiT model as input. The DiT denoiser $\epsilon_\theta(\cdot)$  is trained by: 

\begin{equation}
 \mathcal{L}_{\text{DiT}}=\underset{ \substack{ \boldsymbol{z},    \boldsymbol{\epsilon},  t}}{ \mathbb{E}} \left\|\boldsymbol{\epsilon}-\epsilon_\theta\left(\boldsymbol{z}_t,  t, \boldsymbol{c}_t, \boldsymbol{c}_i \right)\right\|_2^2, \label{s}
\end{equation}
where $\boldsymbol{\epsilon}$ refers to  standard noise and $t$ means denoising timestep. $\boldsymbol{z}_t$ is the video hidden tensor at $t$-th timestep.  

\noindent\textbf{Reward model without using the gradient.} 
In image or video generation,  reinforcement learning (RL) algorithms use rewards to align with preferences. There are two main types: gradient-based rewards (GR) \cite{xu2023imagereward, wu2024deep} and non-gradient-based rewards (NGR) \cite{fan2024reinforcement, black2023training,lee2024parrot}. GR requires the reward gradient for optimization,  whereas NGR does not. The models generate images or videos in pixel space via VAE and then score them using the reward. Backpropagating gradients through the 3D VAEs of most DiT-based video models \cite{yang2024cogvideox, kong2024hunyuanvideo}  is computationally expensive. Therefore,  we choose NGR to optimize the model. The objective function $ L_r$ is defined as maximizing the expected reward,  with its gradient  $\nabla_\theta L_r$ in diffusion models expressed as:

\vspace{-10pt}
\begin{equation}
\nabla_\theta   L_r = {\mathbb{E}} \left[ r(\boldsymbol{x}_0) \sum_{t=1}^T \nabla_\theta \log p_\theta(\boldsymbol{z}_{t-1}|\boldsymbol{z}_t) \right], 
\label{reward}
\end{equation}
where \( \boldsymbol{x}_0 \) donates the generated video in the pixel space and \( \boldsymbol{z}_t \) denotes the latents in the \( t \) -th denoising step. $T$ is the total steps. Omit various control conditions \(\boldsymbol{c}\) for simplicity.

\subsection{Model Architecture}

\subsubsection{SketchDiT}

SketchDiT is designed to efficiently extract the hybrid representation of reference image, sketches and text,  facilitating the implementation of Dynamic Global-Local Memory.

As shown in \cref{fig_model_architecture} (a),  sketches \(\{S_f\}_{f=1}^F\) are first encoded by 3D VAE encoder $\boldsymbol{\mathcal{E}}_v(\cdot)$ to the sketch token \(\boldsymbol{c}_k\). The token is then concatenated with the padded reference image token \(\boldsymbol{c}_i\)  along the channel dimension,  and subsequently concatenated with text features \(\boldsymbol{c}_t\) in the sequence dimension as the input of SketchDiT. We introduce text control conditions in SketchDiT to enable text and reference image to jointly guide animation coloring (\eg,  background generation),  which could not be achieved by past methods. In animation production,  the reference image typically does not always align with the first frame of the sketches. Creators only need to draw a keyframe,  from which smooth videos before and after the key frame can be generated using sketches. Therefore,  during training,  we propose to randomly select the reference image from historical key frame (\eg,  the keyframe that is hundreds of frames away from the current segment.) rather than being fixed as the first frame of the segment to enhance the robustness.

Then,  the concatenated tokens  are fed into the SketchDiT  $\mathcal{S}(\cdot)$ to get the hybrid multimodal feature $\mathcal{S}(\boldsymbol{c}_t, \boldsymbol{c}_i, \boldsymbol{c}_k)$. Following CogvideoX,  SketchDiT adopts \(L\)  DiT modules (where \(L \ll N\) to save computing resources and time consumption),  each of which contains a 3D attention layer and a Feed Forward Network (FFN). To accelerate training,  we use the first $L$ layers of CogvideoX as the initial weights for SketchDiT. The hybrid multimodal features $\mathcal{S}(\boldsymbol{c}_t, \boldsymbol{c}_i, \boldsymbol{c}_k)$ are then injected into the vision branches of different layers' 3D Attention of CogvideoX parallelly. Specifically,  to further reduce overfitting,  we perform skip-layer  control of the base model,  which is as follows: 
\begin{equation}
 \boldsymbol{z}_n=\boldsymbol{z}_n+\gamma \mathcal{S}(\boldsymbol{c}_t, \boldsymbol{c}_i, \boldsymbol{c}_k) \quad  n \in \{2,  4,  \ldots,  N\} , \label{sketch}
\end{equation}
where \(\boldsymbol{z}_n\) refers to the visual features obtained after the 3D Attention in the \(n\)-th DiT block. \(N\) denotes the number of DiT blocks in CogvideoX. $\gamma$ is a weight factor. During training, we freeze the weights of CogVideoX to preserve its original text-guided capabilities.

For the first generation,  the features $\mathcal{S}(\boldsymbol{c}_t, \boldsymbol{c}_i, \boldsymbol{c}_k)$ are directly fed into CogvideoX,  as shown in \cref{fig_model_architecture} (b). For subsequent generation,  the features $\mathcal{S}(\boldsymbol{c}_t, \boldsymbol{c}_i, \boldsymbol{c}_k)$ are fused with relevant global features of Dynamic Global-Local Memory before being fed into CogvideoX,  as shown in \cref{fig_model_architecture} (c).

\subsubsection{Dynamic Global-Local Memory}

Historical animations contain both color consistency features related to the current generation and redundant features. To extract the color consistency features related to the current generation,  we propose the Dynamic Global-Local Memory (DGLM) mechanism as shown in \cref{fig_model_architecture}(c).

The long video understanding (LVU) model Video-XL \cite{shu2024video} is used to extract historical features. After estimating the visual feature changes between adjacent frames using CLIP \cite{wang2024videollamb, radford2021learning},  different segments (\eg,  2,  4,  8 frames,  offering more fine-grained compression than 3D-VAE) are dynamically selected for the video. Subsequently,  these frame segments are fed into the multimodal large language model (MLLM),  which autoregressively generates a visual token \texttt{<vs>} for each segment. Each token \texttt{<vs>} is influenced by all previous frame segments and their \texttt{<vs>} tokens,  while the original frame features are finally offloaded. The key and value of the \texttt{<vs>} token at each layer are stored in the KV cache to improve efficiency. 

Some studies \cite{tao-etal-2024-probing, dang2024explainable} indicate that the middle layers of MLLM focus more on global visual features than the final layer. Specifically,  we define the most recently generated segments as local video \(\mathcal{V}_l\) and all historically generated frames as global video \(\mathcal{V}_g\). Considering these segments comprehensively can achieve better temporal consistency. Specifically,  the global video \(\mathcal{V}_g\) and the local video \(\mathcal{V}_l\) are fed into the LVU model,  which extracts the keys \(\{\boldsymbol{k}_g, \boldsymbol{k}_l\}_m^M\) and values \(\{\boldsymbol{v}_g, \boldsymbol{v}_l\}_m^M\) of the visual token \texttt{<vs>} stored in the KV cache,  where $M$ refers to the extracted layer number. \(\{\boldsymbol{k}_g, \boldsymbol{k}_l\}_m^M\) and \(\{\boldsymbol{v}_g, \boldsymbol{v}_l\}_m^M\) are then fed into the FFN-based projector to align the dimensions of the hybird feature $\mathcal{S}(\boldsymbol{c}_t, \boldsymbol{c}_i, \boldsymbol{c}_k)$. Then the extracted $M$-layer KV cache features and features $\mathcal{S}(\boldsymbol{c}_t, \boldsymbol{c}_i, \boldsymbol{c}_k)$ are sequentially fed into $M$ cross-attention layers,  adaptively extracting global visual features relevant to the current generation as follows:

\vspace{-10pt}
\begin{equation}
\label{eq:attention}
 \text{Softmax}\left( \frac{ \boldsymbol{Q} \cdot \boldsymbol{K}\left( [ \boldsymbol{k}_g^m,  \boldsymbol{k}_l^m] \right)}{\sqrt{d}} \right)  \boldsymbol{V}\left( [ \boldsymbol{v} _g^m,   \boldsymbol{v}_l^m] \right), 
\end{equation}
where the Query $\boldsymbol{Q}=W_q \cdot \mathcal{S}(\boldsymbol{c}_t, \boldsymbol{c}_i, \boldsymbol{c}_k)$,  key  $\boldsymbol{K}=W_k \cdot [ \boldsymbol{k}_g^m,  \boldsymbol{k}_l^m]$ and value $\boldsymbol{V}=W_v \cdot [ \boldsymbol{v}_g^m,  \boldsymbol{v}_l^m]$. $W_q$,  $W_k$ and $W_v$ are weight parameters. \( [ \  ,  \  ] \) denotes feature concatenation.

Summarily,  DGLM achieves long-term color consistency by dynamically compressing historical features and adaptively extracting current-relevant features.

\subsubsection{Color Consistency Reward} 

While the DGLM module significantly improves long-term color consistency,  minor color details remain flawed. Inspired by non-gradient reward studies\cite{fan2024reinforcement, black2023training} in image generation,  we propose the Color Consistency Reward (CCR) to refine the long-term color consistency of animation.

Many past studies \cite{si2022inception, wang2022anti} have shown that self-attention in Transformer acts as low-pass filter,  which means it captures low-frequency features (\eg,  colors in anime) better than high-frequency features (\eg,  sketches). Thus,  by aligning the $M$-layer KV cache features from the LVU model of the generated anime with those of the reference anime,  the generated anime colors can be closer to the real anime colors. The reward function is as follows:

\begin{equation}
r = \sum_{m=1}^{M} \left( \|  \boldsymbol{k}_{\text{ref}}^m - \boldsymbol{k}^m \|_2^2 + \| \boldsymbol{v}_{\text{ref}}^m - \boldsymbol{v}^m \|_2^2\right), 
\end{equation}
Where $\{\boldsymbol{k}_{\text{ref}}^m, \boldsymbol{v}_{\text{ref}}^m\}$ donates the features of the $m$-th KV cache of the reference video,  and $\{\boldsymbol{k}^m, \boldsymbol{v}^m\}$ donates the features of the $m$-th KV cache of the generated video.

The NGR loss from \cref{reward} is used to optimize CCR. Aligning the KV caches of the generated video with the reference video further reduces the gap between inference and training. The KV caches are extracted from reference videos during training,  while these features are extracted from past generated videos during inference. Therefore,  the CCR can further refine long-term color consistency.


\begin{figure}[!t]
\centering
\includegraphics[width=1\linewidth]{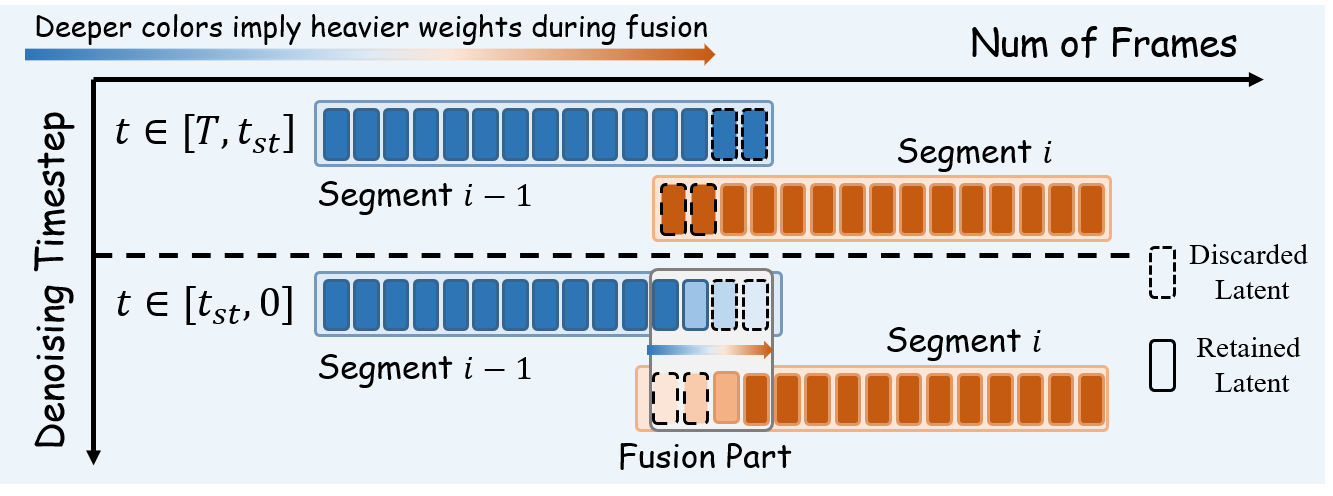}
\caption{Overview of color consistency fusion during inference. The dashed latent of the overlap part is finally discarded.}
\label{lation_fusion}
\end{figure}

\subsection{Inference Paradigm}

While\textit{ LongAnimation} can achieve global consistency between segments,  smooth transitions between adjacent segments also need to be processed. Recent Studies \cite{zhang2024mimicmotion, cai2024ditctrl} explore latent blending strategies in other fields of video generation by blending overlapping segments at all timesteps.

\begin{table*}[tbp]
      \centering
     \resizebox{1\linewidth}{!}
     {
        \begin{tabular}{c c c c c c c c c c c }
            \toprule
             \multirow{2}{*}{ Methods} & \multicolumn{5}{c}{ \textbf{\textit{Short Term} } (14 Frames) } & \multicolumn{5}{c}{ \textbf{\textit{Long Term}} (Average 500 frames) }  \\
            \cmidrule(r){ 2 - 6 } \cmidrule(r){ 7 - 11 } & LPIPS $\downarrow$  & SSIM $\uparrow$ & PSNR $\uparrow$ & FVD  $\downarrow$  &  FID $\downarrow$  & LPIPS $\downarrow$  & SSIM $\uparrow$ & PSNR $\uparrow$ & FVD  $\downarrow$  &  FID $\downarrow$  \\
            \midrule
            ToonCrafter\cite{xing2024tooncrafter} &    0.196&  0.457&   18.57&  564.48&    52.91&  0.238&   0.440&  18.01& 751.20& 89.87\\
            LVCD\cite{huang2024lvcd} &   0.203& 0.732&  22.86&  520.51& 89.39&  0.223& 0.722&  22.77&  734.85&  104.90\\
            AniDoc\cite{meng2024anidoc} &    0.142&  0.759&   24.13& 427.03& 70.31&   0.169&  0.743&   23.17& 531.32& 76.67\\
            LVCD\textsuperscript{*} &    0.126&  0.811&   26.66& 288.70& 78.95& 0.162&  0.776&   22.94& 473.02 & 94.86\\
            \textit{Ours} & \textbf{0.054} 
 & \textbf{0.867}    & \textbf{27.22}   & \textbf{187.48}   & \textbf{37.80} &  \textbf{0.068} & \textbf{0.868} & \textbf{26.71} & \textbf{240.57} & \textbf{40.75} \\
            \midrule
            Improvement &   $\Delta$ 57.1\% &   $\Delta$ 6.9\%  &   $\Delta$ 2.1\%  &  $\Delta$ 35.1\%  &  $\Delta$ 28.6\% &   $\Delta$ \textbf{58.0\%} &  $\Delta$ \textbf{11.8\%} &  $\Delta$ \textbf{15.3\%} &  $\Delta$ \textbf{49.1\%} &  $\Delta$ \textbf{46.9\%}\\
            \bottomrule
            
        \end{tabular}}
    \caption{\textbf{Quantitative comparison} with existing methods. \textit{LongAnimation} achieves the best performance in both short-term and long-term animation coloring. Compared to short-term animation coloring,  \textit{LongAnimation} shows greater improvement in long-term animation coloring. Bolded numbers indicate the best performance.}
    \vspace{-10pt}
\label{main_result}
\end{table*}

\begin{figure*}[ht]
\begin{center}
\includegraphics[width=\textwidth]{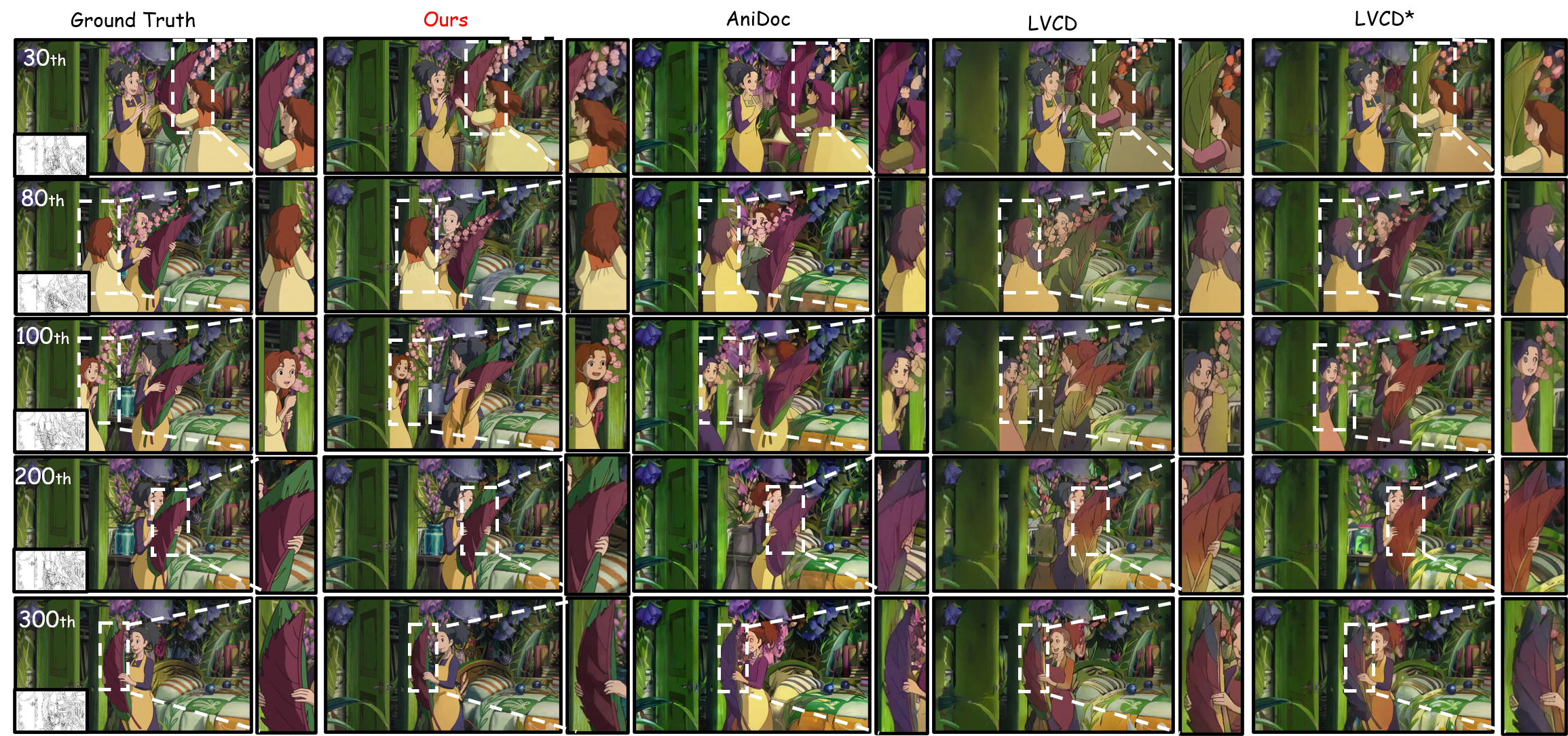}
\end{center}
   \caption{\textbf{Qualitative comparison} with existing methods. \textit{LongAnimation} achieves long-term color consistency through Dynamic Global-Local Memory (\eg,  the gril's dress and leaves). In contrast,  previous methods exhibit unstable color changes. We highly recommend watching the videos in the supplement material for a more direct and clear understanding.}
   \vspace{-10pt}
\label{fig:example}
\end{figure*}

However,  we discover using latent blending strategies for all denoising steps would disrupt visual details (\eg,  brightness),  which is obvious in dark brightness animations,  as shown in \cref{fig_ablation1}. This phenomenon mainly occurs when blending from the early denoising stage. On the one hand,  the early stage focuses on the overall layout features,  while the later stage focuses on the visual details (\eg,  fine-grained color,  brightness). The overall layout features gained by the early denoising fusion are redundant for the animation coloring task. On the other hand,  current DiT-based video models usually jointly decode video latents,  rather than decoding each frame separately like the previous UNet models. It means that unnecessary perturbations to partial latents affect the decoding of all latents. Since visual detail features (\eg,  color,  brightness) are mainly refined in late denoising stage,  we propose a fine-grained color consistency fusion in late denoising stage,  as shown in \cref{lation_fusion}.

Formally,  given two adjacent frame segments $P_{i-1}$ and $P_{i}$,  each containing $F$ frames. The frame number overlapped by two segments is $C$. A fusion factor $\alpha=\frac{1}{C+1}$ is set to control the fusion intensity. $\alpha$ ensures that frames closer to the boundary have lower weights,  while internal frames have higher weights. During denoising,  when $t \in \left[ T,  t_{st} \right]$,  the frame segments are directly concatenated,  with the overlapping part taking half from each segment. When $t \in \left[  t_{st}, 0 \right]$,  latent fusion is applied to the overlapping parts of adjacent segments as follows:

\begin{equation}
\boldsymbol{z}_{it}^k = k \cdot \alpha \cdot \boldsymbol{z}_{it}^k + (1 - k \cdot \alpha) \cdot \boldsymbol{z}_{(i-1)t}^{F-C+k}, 
\end{equation}
where $k\in\left[ 1,  C+1 \right]$ donates the $k$-th overlapping segment frame. $\boldsymbol{z}_{it}^k$ denotes the video latent of the \(k\)-th overlapping frame in segment \(P_i\) during the denoising step \(t\).

In summary,  color consistency fusion smoothes color transitions between adjacent segments by fusing overlap segments during the late denoising stage of inference.

\section{Experiments}
\label{sec:experiments}

\subsection{Experimental Setups}

\textbf{Implementation.} Our model is implemented on CogVideoX-1.5-5B \cite{yang2024cogvideox},  which can stably generate 81 frames. Sakuga-42M \cite{pan2024sakuga} is used as our training dataset. Our work is solely for non-commercial scientific research. We filter high-aesthetic and dynamic video clips with lengths greater than 91 frames,  retaining about 80k videos for training. The model is trained on 6 A100 GPUs with a learning rate 1e-5. The model is trained at a resolution of $1024 \times 576$. We set the SketchDiT layers  \(L = 6\) while  CogVideoX-1.5  layers \(N = 42\). SketchDiT is trained for 30k steps in the first stage,  and GLM is trained for 10k steps in the second stage. Finally,  we use Color Consistency Reward to refine the coloring ability for 10k steps further.

\noindent\textbf{Test Dataset.}  We randomly select 3k samples from the Sakuga testset for short-term coloring testing. To ensure a fair comparison with past models,  we only use the first 14 frames generated by each model for short-term analysis. For long-term generation,  we randomly selected 200 videos, averaging 500 frames to test long-term generation.

 \begin{figure}[!t]
\centering
\includegraphics[width=1\linewidth]{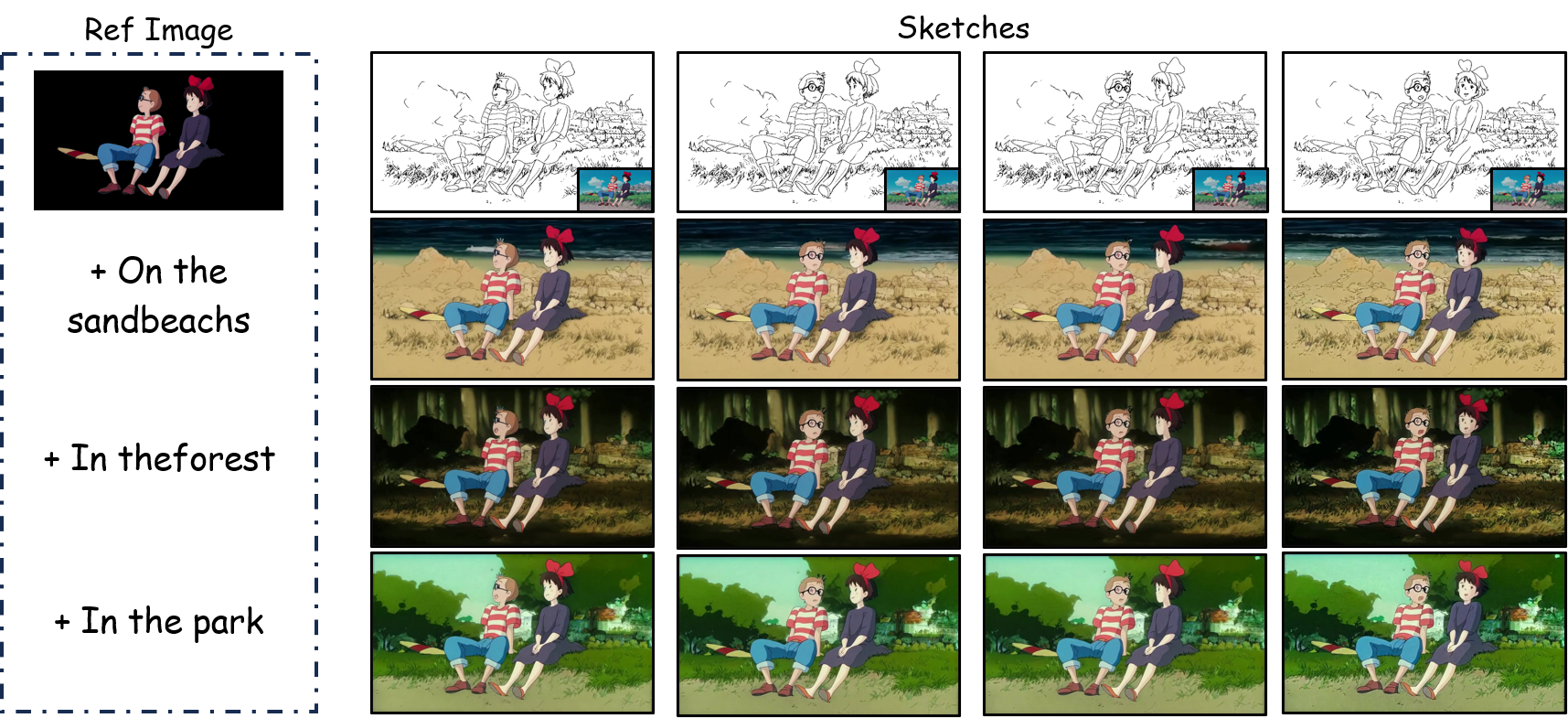}
\caption{Text and reference image jointly control background generation, which could not be achieved by previous methods.}
\vspace{-10pt}
\label{fig_add}
\end{figure}

 \noindent\textbf{Binaryzation.}  Standard animation creation usually uses binary lines. Anidoc \cite{meng2024anidoc} points out that LVCD \cite{huang2024lvcd} directly uses grayscale sketches,  which causes the model to recover the hidden colors from the sketch rather than from the reference image. During training,  we extract the sketches by the method \cite{chan2022learning} and binarize them by setting values greater than 200 to 0 and all other values to 1.

\noindent\textbf{Evaluation metrics.}  We evaluate the animation colorization effect from two aspects: 1) Video quality: FID \cite{heusel2017gans} and FVD \cite{unterthiner2018towards} are used to evaluate the generated frame quality and video quality. 2) Frame colorization similarity. Since the sketches in the animation are extracted from the original animation,  we measure the generated animation frames and the original frames through PSNR,  LPIPS,  and SSIM \cite{wang2004image}. For all metrics,  we resize the frame to $256\times256$ and normalize the pixel to [0, 1] following recent studies \cite{meng2024anidoc, huang2024lvcd}.

 \begin{figure}[!t]
\centering
\includegraphics[width=1\linewidth]{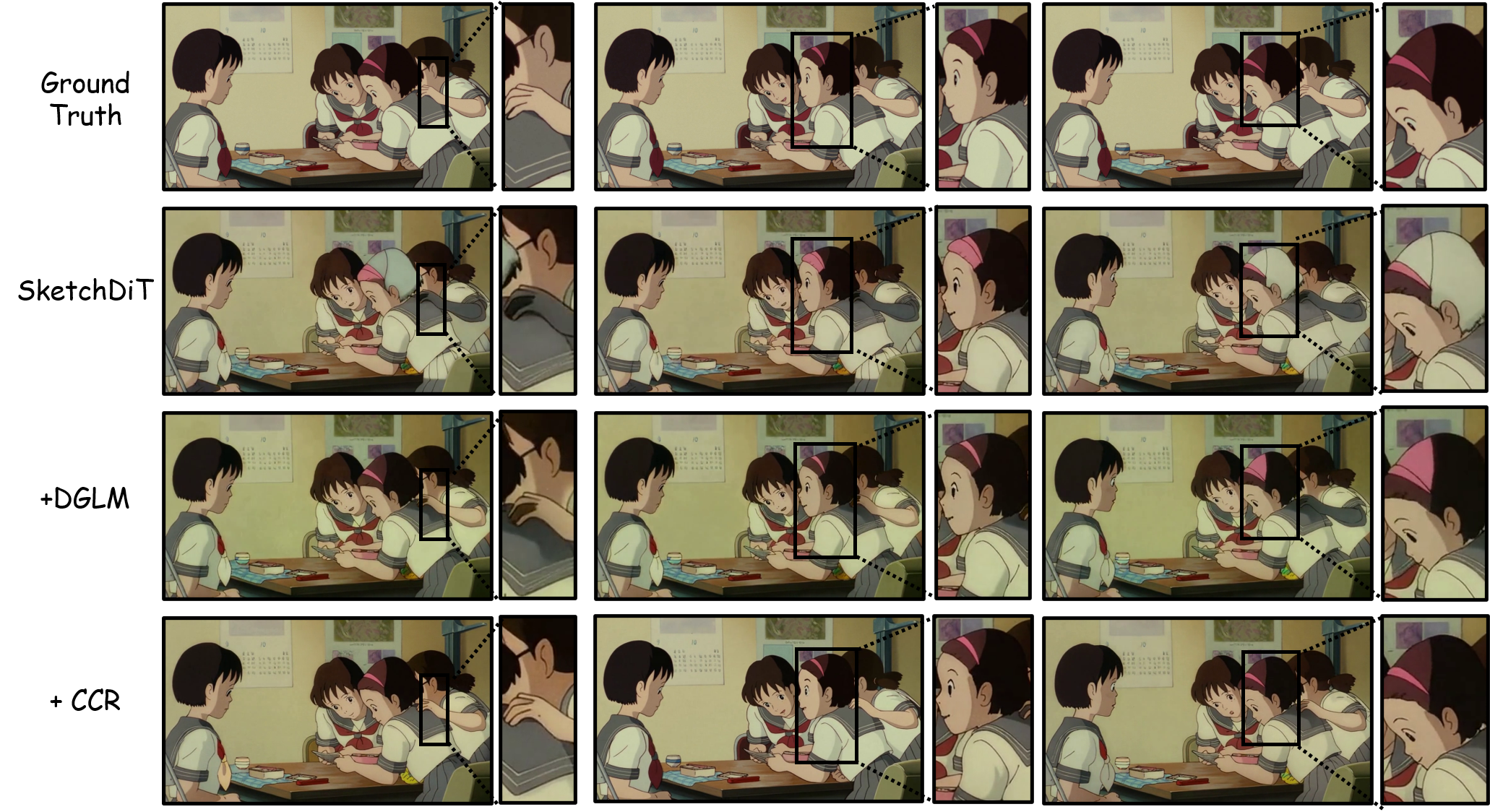}
\caption{\textbf{Ablation Studies} of modules. Compared to SketchDiT,  DGLM markedly enhances color consistency (\eg,  the girl's hair). CCR further refines color details (\eg,  the girl's hairband).}
\label{fig_ablation0}
\end{figure}

\noindent\textbf{Prior SOTAs.} We compare our model with open source state-of-the-arts. These methods are all trained on UNet-based video diffusion models,  which are LVCD \cite{huang2024lvcd},  ToonCrafter \cite{xing2024tooncrafter},  and AniDoc \cite{meng2024anidoc}. Note that in \cref{main_result} and \cref{fig:example},  LVCD\(^*\) \cite{huang2024lvcd} uses grayscale sketches in the [0, 255] range according to the default settings. All other methods use binary sketches. Autoregressively using the last frame of the previous segment as the reference image for the next generation  will lead to the accumulation of noise errors,  as shown in Appendix B.1. Therefore we set all methods to use the same reference frame for generating all segments.

\subsection{Main Results}

In this section,  We will compare \textit{LongAnimation} with prior SOTAs through qualitative and quantitative analyses.

\textbf{Quantitative results.} As shown in \cref{main_result},  our method achieves the best performance in both short-term and long-term animation coloring tasks. For short-term animation generation,  \textit{LongAnimation} improves frame similarity (LPIPS) and video quality (FVD) by 57.1\% and 35.1\% compared to sub-optimal results,  respectively. For long-term video generation,  \textit{LongAnimation} improves frame similarity (LPIPS) and video quality (FVD) by 58.0\% and 49.1\% compared to sub-optimal results,  respectively. \textit{LongAnimation} shows greater improvement in long-term generation compared to short-term generation,  indicating the effectiveness of our model in long-term generation.

\textbf{Qualitative results.} \cref{fig:example} shows the qualitative comparison between \textit{LongAnimation} and existing methods. Our method achieves long-term color consistency through Dynamic Global-Local Memory. More qualitative comparisons are presented in Appendix B.2.

To retain the text-guided ability,  we introduce text features to SketchDiT. As shown in  \cref{fig_add},  \textit{LongAnimation} can achieve long-term text-guided video background generation based on masked reference foregrounds,  which past models \cite{huang2024lvcd, meng2024anidoc, xing2024tooncrafter} can not achieve. \textit{LongAnimation} shows strong generalization and potential for application.

\begin{figure}[!t]
\centering
\includegraphics[width=1\linewidth]{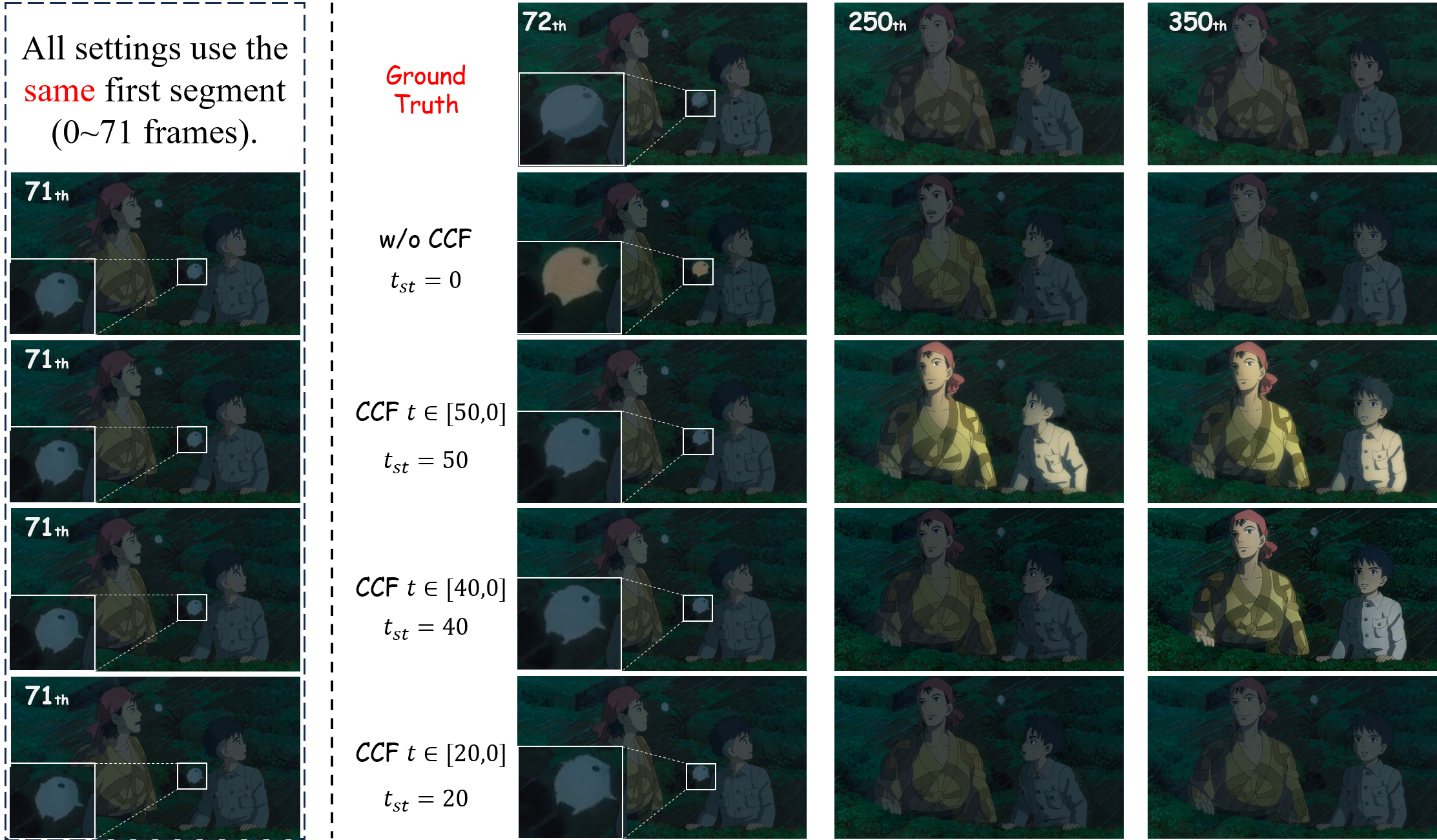}
\caption{\textbf{Ablation Studies} of timestep on color consistency fusion (CCF). Adopting CCF can keep the consistency of the spliced segments (\eg,  white objects in frames 71 and 72). However,  using CCF in the early stage of denoising ($t_{st}=50 \& 40$ ) interferes with other features in latent space, which affects the brightness of other frames (\eg,  frames 250 and 350). The problem does not occur when CCF is used in the late stage of denoising ($t_{st}=20$).}
\vspace{-5pt}
\label{fig_ablation1}
\end{figure}

\begin{table}[!t]
      \centering
     \resizebox{1\linewidth}{!}
     {
        \begin{tabular}{c c c c c c c}
            \toprule
             ID & Settings & LPIPS$\downarrow$  & SSIM $\uparrow$ & PSNR $\uparrow$  & FVD  $\downarrow$  \\
            \midrule
            0 & SketchDiT  & 0.086 & 0.838  &  24.46 & 321.62 &  \\
            1 & 0 + Local Memory & 0.080 & 0.843 &  24.76   & 315.05 \\
            2 & 0 + Global Memory  &  0.078 &  0.854 & 25.38 & 297.93 &  \\
            3 & 0 + DGLM  & 0.076 & 0.862   & 26.11  & 261.76  \\
            4 & 3 + CCR  & \textbf{0.068} & \textbf{0.868}   & \textbf{26.71}  & \textbf{240.57}  \\
            \bottomrule
        \end{tabular}}
    \caption{Ablation studies on each module. Compared with using only SketchDiT (ID-0),  introducing both global memory and local memory (DGLM,  ID-3) significantly improves performance. Compared to local memory (ID-1),  global memory (ID-2) performs better. CCR further refines the coloring performance (ID-4).}
    \vspace{-15pt}
\label{ablation_1}
\end{table}

\begin{table}[!t]
     \centering
      \resizebox{1\linewidth}{!}
      {
        \begin{tabular}{c c c c c c}
        
            \toprule
             Settings  &  LPIPS $\downarrow$  & SSIM $\uparrow$ & PSNR $\uparrow$ & FVD  $\downarrow$    \\
            \midrule
            w/o CCF ($t_{st}=0$) &   0.073 & 0.860 & 26.39  &  266.99  \\
            CCF $t_{st}=50$ &   0.089 & 0.833 &  25.40 & 337.05    \\
            CCF $t_{st}=40$ &  0.083  & 0.850 & 25.71 & 301.62    \\
            CCF $t_{st}=20$ &  \textbf{0.068}   & \textbf{0.868 }& \textbf{26.71}  & \textbf{240.57} \\
            \bottomrule
        \end{tabular}
        }
        
    \caption{Ablation studies for Color Consistency Fusion (CCF),  where \(t_{st}\) represents the timestep for the latent fusion. Injecting local features from the early stage of denoising ($t_{st}=50$ or $t_{st}=40$) reduces the video quality.}
    \vspace{-10pt}
\label{ablation_3}
\end{table}

\subsection{Ablation Studies}
To demonstrate the effectiveness of essential components,  we conduct extensive ablation experiments. All ablation experiments are tested on long video (average 500 frames). We present the key ablation study in the main text,  with the remaining ablation experiments shown in Appendix B.3. 

\textbf{Effectiveness of Each Module for \textit{LongAnimation}}.   As indicated in \cref{ablation_1} and \cref{fig_ablation0},  compared to using only SketchDiT (ID-0),  introducing DGLM (ID-3) significantly improves long-term color consistency,  with the frame similarity metric (LPIPS) improving by 11.6\% and the video quality metric (FVD) improving by 18.6\%. ID-1 and ID-2 validate the effectiveness of global video and local video respectively. Compared to ID-3,  Color Consistency Reward (CCR) (ID-4) further refines the long-term coloring effect,  with frame similarity metric (LPIPS) improving by 10.5\% and video quality metric (FVD) improving by 8.0\%.

\textbf{Effectiveness of Color Consistency Fusion (CCF)}.  As indicated in \cref{ablation_3} and \cref{fig_ablation1},  latent fusion from the early denoising stage ($t_{st}=50 \& 40$) results in a decrease in frame similarity and video quality metrics compared to not using CCF. However,  latent fusion from the late denoising stage ($t_{st}=20$) leads to improvements in these metrics. This indicates that CCF from the late denoising stage not only provides better color consistency in the fusion frames but also maintains the brightness in other frames.

\subsection{Frequency Analysis}

\begin{figure}[!t]
\centering
\includegraphics[width=1\linewidth]{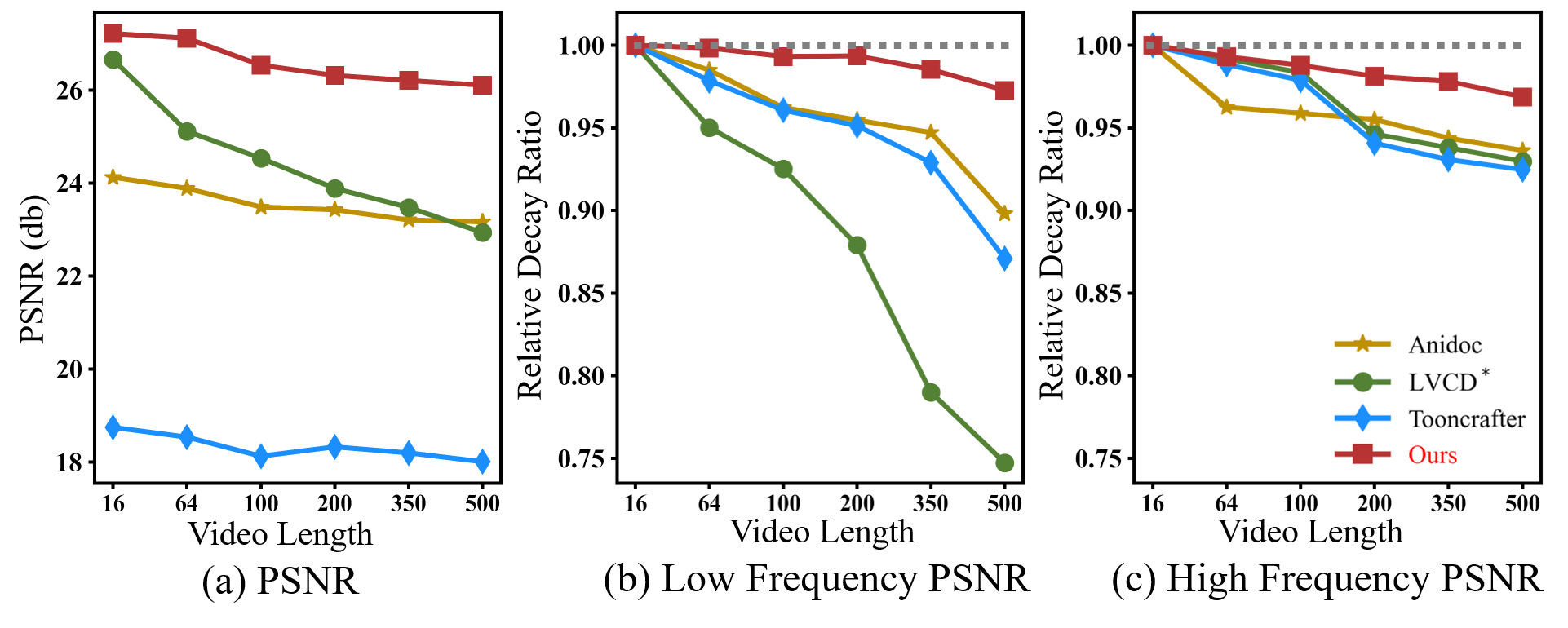}
\caption{(a) PSNR with existing methods. \textit{LongAnimation} outperforms previous methods in PSNR metric. (b) PSNR relative decay ratio in the low-frequency domain. Our method exhibits the least attenuation in low-frequency information (\eg,  color),  indicating that our proposed Dynamic Global-Local Memory can better maintain long-term color consistency. (c) PSNR relative decay ratio in the high-frequency domain. Our method exhibits the least attenuation in high-frequency information (\eg,  sketches). Since all methods are controlled by sketches,  their attenuation in high frequencies is generally smaller compared to low frequencies.}
\vspace{-10pt}
\label{visualization_1}
\end{figure}

As shown in \cref{visualization_1} (a),  \textit{LongAnimation} has the best PSNR performance. To further demonstrate the superiority of \textit{LongAnimation},  we perform frequency domain analysis by applying the Fourier transform to the videos generated by each model and reference videos. After separating high-frequency (\eg,  sketch) and low-frequency (\eg,  color) features,  we calculate the PSNR for each frequency component. For a fair comparison,  we calculate the PSNR degradation rate between long videos and short videos (\emph{i.e.},  14 frames) generated by each model. As shown in \cref{visualization_1} (b),  our method has the weakest decay ratio at low frequencies,  improved by 8.2\% at 500 frames compared to sub-optimal method. It shows that our method better preserves the low-frequency (i.e.,  color) features,  while the low-frequency preservation of past methods diminishes over time. Since all methods use high-frequency sketch control,  high-frequency decay is generally small,  as shown in \cref{visualization_1} (c).

\section{Conclusion}
In this paper,  we propose \textit{LongAnimation},  a novel long animation coloring framework to achieve long-term color consistency. \textit{LongAnimation} dynamically compresses historical features and adaptively extracts features relevant to current generation by a Dynamic Global-Local Memory mechanism. Color consistency is further refined by Color Consistency Reward. To support DGLM,  we propose SketchDiT to extract hybrid reference features. Extensive experiments demonstrate the superiority of our method,  showing strong application prospects in the animation industry.

\section{Acknowledgement}
This research is supported by Artifcial IntelligenceNational Science and Technology Major Project 2023ZD0121200, National Natural Science Foundation of China under Grant 62222212 and 623B2094.

{
    \small
    \bibliographystyle{ieeenat_fullname}
    \bibliography{main}
}

\end{document}